# Unsupervised Identification and Replay-based Detection (UIRD) for New Category Anomaly Detection in ECG Signal

Authors: Zhangyue Shi[1], Zekai Wang[2], Yuxuan Li[3+]

[1]Independent Researcher, St. Louis, MO, United States,

[2]Charles F. Dolan School of Business, Fairfield University, Fairfield, CT, USA United States,

[3]School of Business, East China University of Science and Technology, Shanghai, China

[*] Zhangyue Shi and Zekai Wang contributed equally to this work

[+]Corresponding author yuxuan.li@ecust.edu.cn

**Abstract**: In clinical practice, automatic analysis of electrocardiogram (ECG) is widely applied to identify irregular heart rhythms and other electrical anomalies of the heart, enabling timely intervention and potentially improving clinical outcomes. However, due to the limited samples in certain types of ECG signals, the class imbalance issues pose a challenge for ECG-based detection. In addition, as the volume of patient data grows, long-term storage of all historical data becomes increasingly burdensome as training samples to recognize new patterns and classify existing ECG signals accurately. Therefore, to enhance the performance of anomaly detection while addressing storage limitations, we propose a pseudo-replay based semi-supervised continual learning framework, which consists of two components: unsupervised identification and replay-based detection. For unsupervised identification, an unsupervised generative adversarial network (GAN)-based framework is integrated to detect novel patterns. Besides, instead of directly storing all historical data, a pseudo replay-based learning strategy is proposed which utilizes a generator to learn the data distribution for each individual task. When a new task arises, the generator synthesizes pseudo data representative of previous learnt classes, enabling the model to detect both the existed patterns and the newly presented anomalies. The effectiveness of the proposed framework is validated in four public ECG datasets, which leverages supervised classification problems for anomaly detection. The experimental results show that the developed approach is very promising in identifying novel anomalies while maintaining good performance on detecting existing ECG signals.

# 1 Introduction

## 1.1 Research Background

Cardiovascular Diseases (CVDs) remain the leading cause of death worldwide, accounting for roughly 17.9 million deaths annually [1]. Early detection and management of heart abnormalities are therefore critical. The electrocardiogram (ECG) is a primary, non-invasive tool for monitoring the heart's electrical activity and is essential in diagnosing cardiac conditions such as coronary artery disease, myocardial infarction, and especially cardiac arrhythmias [2]. By capturing the electrical signals associated with each heartbeat, the ECG provides clinicians with real-time insight into heart rhythm and conduction. In clinical practice, ECG is the benchmark for identifying irregular heart rhythms and other electrical anomalies of the heart. Hence, automatic analysis of ECG data can facilitate early detection of arrhythmias like atrial fibrillation or ventricular tachycardia, saving valuable time for clinicians in diagnosis and treatment planning. Moreover, it supports continuous, long-term monitoring of patients with chronic cardiac conditions, which is crucial for preventing adverse events and improving overall disease management.

ECG signals are time-series data with rich temporal dynamics: each heartbeat produces a characteristic waveform (P-QRS-T complex) reflecting the cardiac cycle. ECG data presents several properties. First, the nonlinear and nonstationary nature of cardiac dynamics introduces significant inter-patient variability – the morphology of ECG waveforms can differ substantially between individuals due to anatomical, physiological, and pathological differences. As a result, an algorithm trained on one cohort may fail to generalize to others, since what constitutes a "normal" ECG pattern can vary widely across the population. In addition, intra-patient variability further complicates the problem: even within the same individual, ECG characteristics can shift over time due to changes in heart rate, activity levels, autonomic regulation, and evolving health conditions [3]. These temporal variations reflect the inherently dynamic behavior of the cardiovascular system and challenge the models which assume stationary input distributions. Meanwhile, ECG recordings are frequently contaminated by various types of noise and artifacts, including baseline wander, muscle (EMG) noise, powerline interference, and electrode motion artifacts [4]. These unpredictable disturbances further amplify the nonlinear and nonstationary properties of the signal, making it difficult to consistently capture subtle arrhythmic patterns. As a result, models must be capable of learning from complex, evolving, and noisy data environments to ensure reliable anomaly detection in ECG monitoring systems. In addition, many pathological patterns (arrhythmias) are episodic or rare. A standard 10-second or 30-minute ECG may miss transient arrhythmias if they do not occur during the recording. Therefore, when building datasets, normal beats vastly outnumber abnormal ones, leading to severe class imbalance issues [5]. The class imbalance issues pose a challenge for traditional supervised learning, which can be biased toward the majority class and struggle to learn rare event patterns.

The aforementioned ECG signal properties highlight a fundamental limitation in traditional supervised learning approaches: they often assume that sufficient and fully representative data are available before model training. However, this assumption rarely holds in real-world healthcare systems, particularly for ECG signal analysis. In practice, new types of arrhythmic patterns can emerge unpredictably, and data are collected incrementally over time as patients' conditions evolve. For example, in ECG monitoring, different types of anomalies appear across time (see Figure 1), each reflecting distinct and previously unseen patterns in sensor signals. These variations correspond to shifts in the underlying data distribution, making it essential for the system to detect and adapt to novel anomalies. Unsupervised learning frameworks are particularly suited for detecting such novel anomalies without requiring explicit labels. By learning the distribution of normal ECG signals, these models can flag deviations as potential anomalies, addressing the scarcity of labeled abnormal data. However, once a novel anomaly type is detected, retraining the model becomes necessary to recognize and classify this new pattern accurately. A major challenge in retraining is catastrophic forgetting - when a model, updated with only new data, suffers from a sudden degradation in performance on previously learned tasks [6]. A naive solution might involve retraining the model from scratch each time a new arrhythmic pattern arises. Unfortunately, this approach may be impractical in real-world scenarios: As the volume of patient data grows, long-term storage of all historical data becomes increasingly infeasible due to hardware limitations, privacy regulations, or data retention policies. Despite these challenges, the need to retrain a model that has good overall performance on both previously learnt patterns and new novelty remains critical. Consequently, there is an urgent need for a methodology that enables efficient incremental training of machine learning models in healthcare systems.

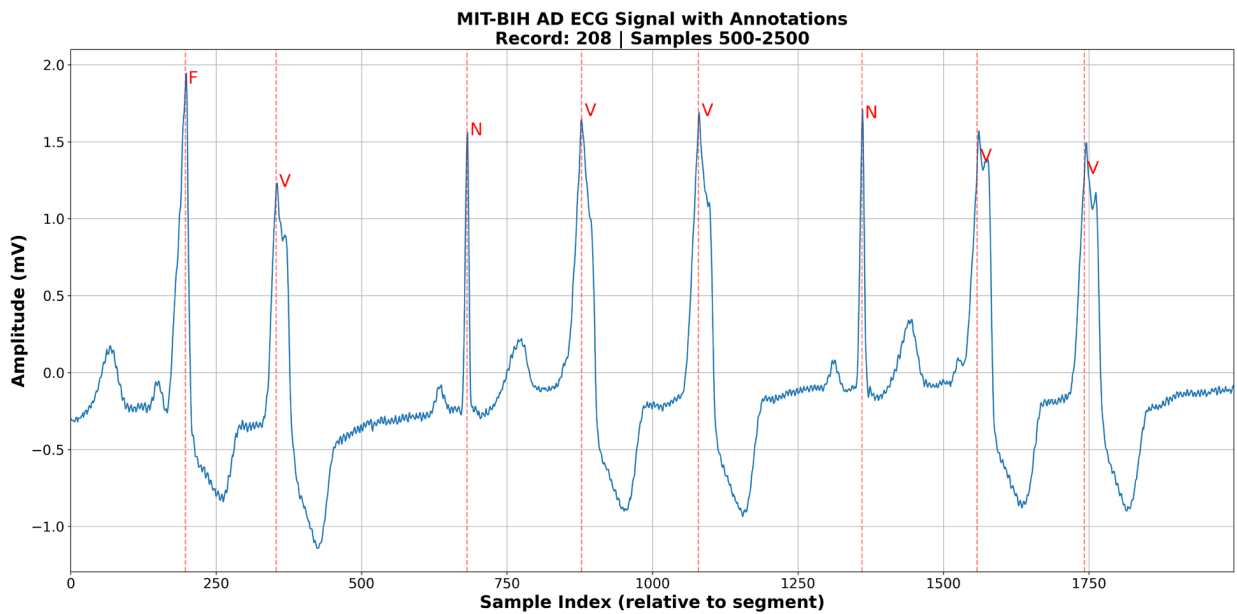

*Figure 1. **ECG segment from the MIT-BIH Arrhythmia Database (Record 208, Samples 500–2500).** The plot shows a sequence of cardiac beats, including normal beats (N), fusion beats (F) and premature ventricular contractions (PVCs labeled as V). Fusion beats arise from the simultaneous activation of the ventricles by both normal atrial impulses and ectopic ventricular impulses, resulting in a hybrid QRS morphology. PVCs are characterized by wide and abnormal QRS complexes without preceding P-waves. Normal beats (N) are also labeled for comparison. Red dashed lines indicate the annotated locations and it is clear to see that various signal patterns may occur at different time.*

## 1.2    Research Objective

To enhance the performance of anomaly detection while addressing storage limitations, we propose an unsupervised identification and replay-based detection (UIRD) framework. This framework consists of two components: novel anomaly detection and continual learning. For anomaly detection, an unsupervised generative adversarial network (GAN) is developed to detect novel patterns. Then it comes to model retraining. Instead of directly storing all historical data, a pseudo-replay based learning strategy is proposed which utilizes a generator to learn the data distribution for each individual pattern. When a new anomaly arises, the generator synthesizes pseudo data representative of previous learnt patterns, enabling the model to be trained on both the newly generated pseudo data and the data from the current abnormal pattern. Compared to existing class-incremental continual learning methods, the proposed framework offers several key advantages:

1. Memory-augmented Autoencoder with Generative Adversarial Network (MadeGAN) for novel class detection: MadeGAN introduces a memory-augmented deep autoencoder combined with adversarial training to enhance ECG anomaly detection. By explicitly storing and retrieving diverse normal patterns through a memory module, it enables more accurate reconstruction of normal signals and improves sensitivity to anomalies without relying on large labeled datasets. Additionally, adversarial training with a feature-matching strategy enriches the learned normal data distribution, making the model more robust against noise and subtle signal variations. Compared to traditional autoencoders and GAN-based anomaly detectors, MadeGAN addresses the common issues of overfitting, poor generalization, and training instability, resulting in more reliable and robust anomaly detection in ECG analysis.
2. Integration of Anomaly Detection and Continual Learning: Many existing continual learning approaches assume prior knowledge of when novel anomalies occur, which is often impractical in real-world scenarios. These methods lack a clear mechanism to determine when to initiate continual learning. To address this limitation, our study integrates anomaly detection and continual learning into a unified framework. This integration creates a closed-loop system that seamlessly connects anomaly detection with model retraining, ensuring timely and adaptive updates in response to newly identified anomalies.

3. Performance and Hardware Efficiency: The pseudo replay-based learning framework strikes an effective balance between maintaining high model performance and reducing hardware constraints, making it suitable for resource-limited environments.

The rest of the structure of this paper is organized as follows. In Sec 2, a brief literature review is provided. Then the proposed research methodology framework is elaborated in Sec 3. To validate the effectiveness of the proposed method, Sec 4 presents four real-world case studies. Eventually, conclusions are summarized, and future works are discussed in Sec. 5.

## 2 Literature review and research background

Since this work desires to not only identify unsupervised anomalies but also detect existing anomalies with pseudo data, the literature review related to anomaly detection in ECG signals is illustrated in Sec. 2.1 and Sec. 2.2 reviewed the literatures related to continual learning-based applications. Afterwards, the research gaps are summarized in Sec. 2.3.

### 2.1 Anomaly Detection in ECG Signal

Early approaches to automatic ECG anomaly detection employ shallow supervised classifiers such as support vector machines (SVMs), decision trees, and k-nearest neighbors ($k$-NN). These models require labor-intensive manual feature engineering – for example, extracting wavelet coefficients, heartbeat intervals, and morphological descriptors of P-QRS-T complexes – to characterize the ECG signal [7], [8]. Designing these features is not only highly time-consuming but also requires significant domain expertise in signal processing and cardiology, making it difficult to scale or automate the model development process. Furthermore, despite careful feature design, traditional methods achieve only moderate accuracy and often struggle to generalize. Additionally, the optimal feature set for one dataset often does not transfer well to others, reflecting poor generalization to unseen data and patient populations.

The advent of deep learning brings significant improvements in ECG anomaly detection through models capable of automatic feature extraction. Researchers applied convolutional neural networks (CNNs), recurrent neural networks (RNNs), and other deep architectures to learn hierarchical features directly from raw ECG signals [9], [10]. This eliminates the need for manual feature selection and yields more informative representations, resulting in higher classification accuracy than earlier methods. For example, a deep CNN model trained on large ECG datasets is able to classify 12 cardiac rhythm classes with expert-level accuracy [11]. Nevertheless, purely supervised deep learning frameworks have notable limitations. They require large volumes of annotated ECG data for training and often suffer from extreme class imbalance in arrhythmia datasets (e.g., predominance of normal sinus rhythm over rare arrhythmias). Performance can degrade significantly when the data are imbalanced.

To address the shortcomings of supervised schemes, recent research has focused on unsupervised anomaly detection methods that learn the distribution of normal ECG patterns and flag deviations as potential anomalies. Approaches based on autoencoders or generative adversarial networks (GANs) are typically trained using only normal ECG data [12]. During deployment, an input that cannot be well reconstructed or generated is deemed anomalous, since it falls outside the learned "normal" manifold. Such frameworks can detect arbitrary abnormal rhythms without explicit labels for each arrhythmia type. However, standard autoencoders tend to overfit by essentially memorizing the training set, leading to identity-mapping and failure to generalize to slight distribution shifts or noise. Likewise, GAN-based anomaly detectors (e.g., AnoGAN [13] and its variants) have been prone to training instability and poor control, which undermines their reliability on ECG data. Under such circumstances, how to identify anomalies while detecting existing anomalies remains challenging. To achieve this, continual learning-based approaches need to be applied and the related literatures are reviewed in Sec. 2.2.

## 2.2 Continual learning and its application in Physiological Signal Data

Traditional machine learning models operate under the assumption that training and test data are drawn from the same distribution, which forms the foundation for the model's ability to generalize to unseen data. However, this assumption often fails to hold in real-world systems, where training data for a specific learning task may only be available during limited time periods. In such scenarios, the emergence of novel data classes necessitates the development of new models. Continual learning addresses this challenge by enabling the incremental acquisition of new skills without degrading previously learned knowledge [14]. This approach effectively balances the stability and flexibility of machine learning models, making it well-suited for dynamic environments.

State-of-the-art continual learning methods can be broadly categorized into several groups: memory-based replay methods, parameter regularization methods, functional regularization methods, dynamic networks, knowledge distillation, and model rectify [14], [15]. Memory-based replay methods typically employ a fixed-size memory buffer to store samples from previously encountered classes [16] or adopt a pseudo-replay strategy to generate synthetic input data or latent features [6], [17]. Parameter regularization methods, such as Elastic Weight Consolidation (EWC), Memory-Aware Synapses (MAS), and Synaptic Intelligence, introduce a regularization term into the loss function to penalize significant changes to parameters critical for previously learned tasks [18], [19], [20]. Similarly, functional regularization methods incorporate a regularization term into the loss function, but their focus is on preserving the input-output mapping across tasks. Notable examples include Learning without Forgetting (LwF) and Functional-Regularization of Memorable Past (FROMP) [21], [22]. Dynamic networks based method expands the network structure dynamically to accommodate novel classes with technique such as adding neurons, backbones, or prompts to increase model capacity while not overwrite the old features [23], [24]. Knowledge distillation transfers knowledge from old teacher model to new student model across different stages [25]. Model rectify tries to rectify the biases in the model caused by incremental learning. Common model rectify methods include unified classifier incrementally via rebalancing (UCIR) and weight aligning [26], [27].

Among these approaches, memory-based replay methods have demonstrated superior performance in class-incremental continual learning scenarios [14], [28]. There are already several applications of continual learning in healthcare, to be more specific, physiological signal analysis. In terms of memory-replayed methods, Kiyasseh *et al.* proposed a replay-based method for cardiac arrhythmia classification [29]. Speaking of parameter regularization methods, Aslam *et. al* applied EWC to make predictions for different diseases outbreak [30]. For the application of functional regularization methods, Ammour *et al.* utilized LwF for heartbeat classification [31]. Sun *et al.* developed a novel self-attention mechanism to learn incrementally by calculating the non-overlapping loss in medical time series data [32].

While continual learning holds a significant promise for anomaly detection, several limitations still remain. In the case of parameter regularization and functional regularization methods, the model architecture is inherently constrained, as these approaches rely on calculating regularization losses based on pre-existing neurons. This rigidity restricts the flexibility to adapt the model structure to evolving data patterns, limiting their effectiveness in dynamic environments. Additionally, these methods perform optimally when applied to tasks with similar characteristics. However, in ECG signal analysis, anomalies often exhibit diverse and dissimilar patterns, which can undermine the efficacy of monitoring systems. For class-incremental tasks, memory-based continual learning methods have been shown to achieve superior performance [14]. As data volumes grow, the long-term retention of process data becomes increasingly burdensome due to hardware limitations, posing a significant challenge for scalability especially for the large data scale modeling tasks.

## 2.3 Research Gaps

The research gaps in class-incremental continual learning methods are multifaceted. First, as far as authors know, there is no existing framework that integrates both anomaly detection and continual learning together. Continual learning work assumes the occurrence of new class is known and then they can directly work on the continual model training. However, it may not be true in real-world cases since when and how new class of anomaly occurs is not pre-determined by humans. Therefore, it is of great significance to develop a framework which can detect the occurrence for new class of anomalies and conduct continual learning together.

In terms of unsupervised anomaly detection, traditional machine learning methods need careful feature engineering and their model generalization ability is poor. For supervised neural network-based methods, they often require large volumes of annotated data for training and often suffer from data imbalance.

Meanwhile, unsupervised anomaly detection methods can get overfitting easily by essentially memorizing the training set and fail to generalize to slight distribution shifts or noise.

Also, speaking of current continual learning approaches, while memory-replay methods based on data generation provide a promising solution by synthesizing data that closely resembles that of previous tasks and reducing storage requirements, their application in ECG signal analysis remains underexplored [29]. Additionally, existing replay-based continual learning methods face challenges in achieving high-accuracy detection performance due to hardware limitations. Regarding regularization-based continual learning methods, their inflexibility in model architecture poses a significant constraint. These methods require the use of the same model architecture for both previous and new tasks, which limits the ability to select the most suitable model based on the specific characteristics of the data. Furthermore, current continual learning approaches often overlook the potential for improving model performance through enhanced data quality. To address these gaps, Sec. 3 introduces a framework consisting of anomaly detection and pseudo-replay based continual learning, designed to overcome these limitations.

## 3 Research methodology

### 3.1 Research framework of overall methodology

This section presents an unsupervised identification and replay-based detection (UIRD) framework that leverages both Memory-Augmented Deep Autoencoder with Generative Adversarial Networks (MadeGAN) and pseudo replay-based data generation. As illustrated in Figure 2, the framework consists of two key stages: (1) novel anomaly detection (seeing Figure 2 (a)); and (2) continual learning (seeing Figure 2 (b) and (c)). First, a novel autoencoder integrated with a memory module, trained within a GAN-based framework is employed [5], consisting of differentiating novel ECG signal type from existing ECG types and handling data-lacking and imbalanced issues. If a novel group of anomalies is detected, then continual learning will be triggered. In the pseudo-replay based continual learning framework, a generator learns the data distribution of each class as it appears. Multiple generative methods - such as the Synthetic Minority

Oversampling Technique (SMOTE) and Generative Adversarial Networks (GANs) [33], can be employed due to their ability to capture underlying distributions and produce synthetic data. The selected generator (validated for effectiveness) creates pseudo data for existing classes without requiring full dataset storage. When a new anomaly is introduced, these pre-trained generators synthesize high-quality data representing prior classes, enabling model updates using both synthetic and newly collected data. In this work, SMOTE is adopted for its superior performance in data augmentation (see Sec. 4 for details). Sec. 3.2 provides an overview of MadeGAN, while Section 3.3 elaborates on the SMOTE and Section 3.4 illustrates the paradigm of the proposed framework.

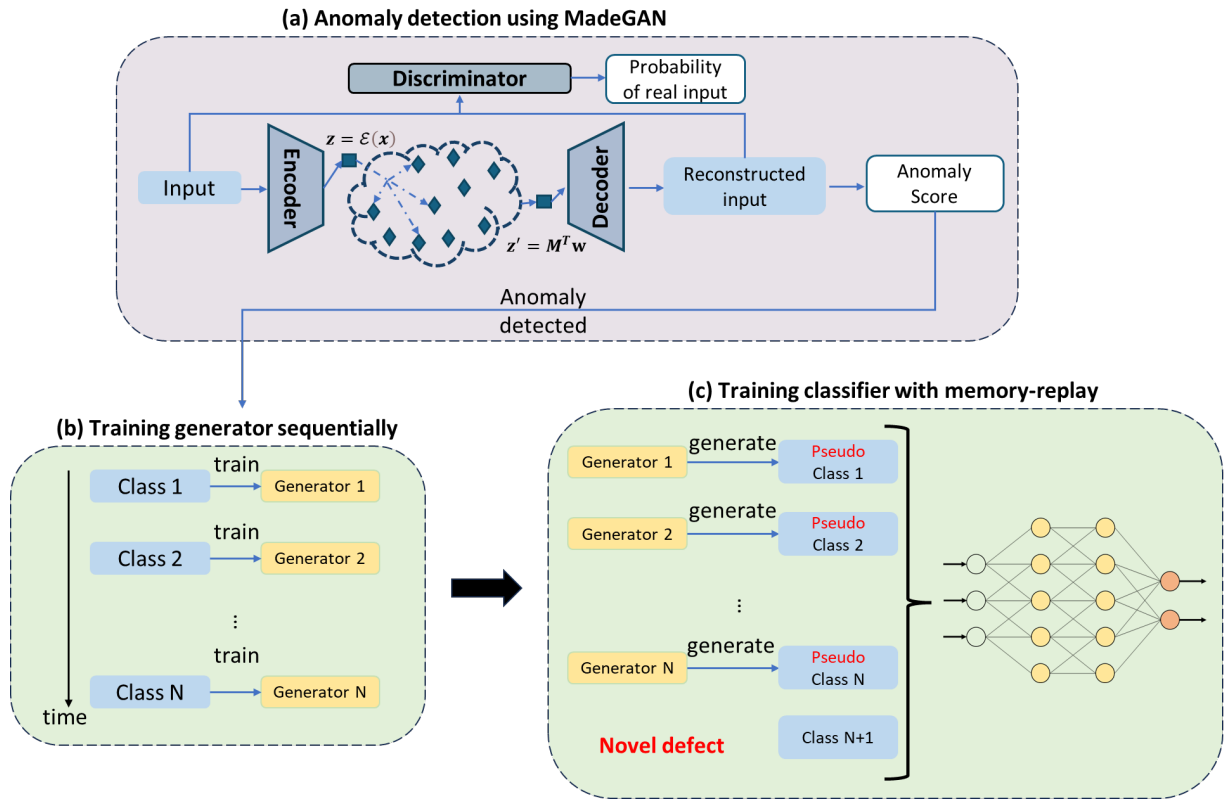

*Figure 2. (a) Anomaly detection with MadeGAN: Detect if novel anomaly has occurred using reconstruction error from MadeGAN, moving to continual learning part if novel anomaly detected. (b) Sequential generator training: Each generator is trained independently in its corresponding class to learn the underlying data distribution. (c) Model update with novel anomalies: When a new anomaly emerges, the pre-trained generators synthesize data for existing classes. This synthetic data,*

*combined with the new anomaly samples, enables the model to update its discriminative capability across both historical and novel classes.*

## 3.2 Unsupervised anomaly detection for new class identification

In this study, we employ MadeGAN model [5] to do anomaly detection for new class identification. MadeGAN is a framework combining a memory-augmented autoencoder (MemAE) for anomaly detection with adversarial training. Specifically, as shown in Figure 2(a), the MemAE consists of three key components: 1). the encoder $\mathcal{E}$ maps an input signal $\boldsymbol{x} \in \mathbb{R}^{d_x \times 1}$ to a latent query $\boldsymbol{z} = \mathcal{E}(\boldsymbol{x}) \in \mathbb{R}^{d_z \times 1}$; 2). A memory module stores $K$ trainable prototype vectors of normal ECG features in a matrix $\boldsymbol{M} \in \mathbb{R}^{K \times d_z}$; it computes cosine-similarity weights $w_i$ between $\boldsymbol{z}$ and each prototype $\boldsymbol{m}_i$ in $\boldsymbol{M}$ and returns a retrieved latent $\boldsymbol{z'} = \boldsymbol{M}^T \boldsymbol{w}$, where $\boldsymbol{w} \in \mathbb{R}^{K \times 1}$; 3). The decoder $\mathcal{D}$ then reconstructs $\hat{\boldsymbol{x}} = \mathcal{D}(\boldsymbol{z'})$. By training on normal data with reconstruction loss $\mathcal{L}_{rec} = \|\boldsymbol{x} - \hat{\boldsymbol{x}}\|^2$, the model ensures that normal signals are reconstructed well (low error) while anomalies produce large errors. We define the anomaly score $S(\boldsymbol{x}) = \|\boldsymbol{x} - \mathcal{D}(\boldsymbol{z'})\|$, so that a high $S(\boldsymbol{x})$ flags an ECG cycle as abnormal.

Training the MemAE alone yields a reconstruction-based detector, but to enrich the learned normal patterns and improve robustness, adversarial training is intergrated. In MadeGAN, the MemAE (encoder $\mathcal{E}$ + memory module $\boldsymbol{M}$ + decoder $\mathcal{D}$) is treated as the GAN generator $\mathcal{G}$ and a 1D-CNN discriminator $\mathcal{F}$ is added. The discriminator is trained to distinguish real normal ECGs from the reconstructed (generated) signals.

To train the MadeGAN framework, a composite loss is employed. The generator, MemAE loss include:

- Reconstruction loss $\mathcal{L}_{rec} = \|\boldsymbol{x} - \hat{\boldsymbol{x}}\|^2$ to make reconstructions faithful;
- Feature-matching loss $\mathcal{L}_{fm} = \|h_\mathcal{F}(\boldsymbol{x}) - h_\mathcal{F}(\hat{\boldsymbol{x}})\|^2$, matching intermediate features $h_\mathcal{F}(\cdot)$ of discriminator $\mathcal{F}$ for real vs. fake signals. This encourages encoder $\mathcal{E}$ to preserve realistic ECG features;

- Sparsity loss $\mathcal{L}_{sp} = \|w\|_1$ on the memory weights, which forces the memory to use a few strong prototypes and improves interpretability.

Meanwhile the discriminator loss is the usual GAN loss: $\mathcal{L}_{GAN} = \log \mathcal{F}(x) + \log(1 - \mathcal{F}(\hat{x}))$. We alternate updates so that $\mathcal{F}$ learns to assign high probability to real signals and low to reconstructions, while $\mathcal{E}$ (MemAE) tries to fool $\mathcal{F}$ in addition to minimizing the loss function of MemAE, $\mathcal{L}_G = \mathcal{L}_{rec} + \mathcal{L}_{fm} + \mathcal{L}_{sp}$. As such, the overall objective of MadeGAN is given by

$$\min_{\theta_G} \max_{\theta_\mathcal{F}} \mathcal{L}_{total}(\theta_G; \theta_\mathcal{F}) = \mathcal{L}_{GAN}(\theta_G; \theta_\mathcal{F}) + \mathcal{L}_G(\theta_G; \theta_\mathcal{F}) \tag{1}$$

This adversarial augmentation has the effect of generating diverse, realistic ECG patterns and enriching the normal data distribution, making the anomaly detector more robust (reconstructions improve for normal signals and remain poor for anomalies).

During inference, an unseen ECG cycle $x'$ is processed by the well-trained MadeGAN to compute an anomaly score $S(x')$ and detect if it is a novel class. Formally, the inference steps are summarized in Algorithm 1.

**Algorithm 1** MadeGAN Inference

---

**Input**: Well-trained encoder $\mathcal{E}$, memory module $M$, and decoder $\mathcal{D}$; threshold $\tau$; new ECG cycle $x'$.
**Step 1:** $z' = \mathcal{E}(x')$
**Step 2:** Compute $w[i] = cosine\_similarity(z', M[i])$
**Step 3:** $z'' = M^T w$
**Step 4:** $\hat{x}' = \mathcal{D}(z'')$
**Step 5:** Anomaly score $= S(x') = \|x' - \mathcal{D}(z'')\|^2$
**Step 6:**
    **If** anomaly score $> \tau$:
        Label = "novel class"
    **Else**:
        Label = "existing class"
    **End if**
**Output:** Anomaly score; Label

## 3.3 SMOTE for pseudo ECG signal generation

This section outlines the methodology for synthesizing high-quality artificial data to augment pre-existing class representations through the application of the Synthetic Minority Oversampling Technique (SMOTE). SMOTE, initially developed to mitigate class imbalance - a scenario characterized by disproportionate representation between minority and majority class samples - operates by generating synthetic instances for underrepresented classes. The technique achieves this by interpolating new data points along the feature space vectors connecting existing minority class samples to their $k$ nearest neighbors, where $k$ represents a user-defined hyperparameter.

As depicted in Figure 3, for a hypothetical configuration with $k = 3$, a minority class instance $x_1$ is paired with its three nearest neighbors $(x_2, x_3, x_4)$ within the minority class. Synthetic samples (denoted $a, b, c$) are generated by randomly selecting positions along the linear segments between $x_1$ and each neighbor. The total number of synthetic instances produced is proportional to the product of $k$ and the original minority sample count. When the required augmentation exceeds this product, all interpolated points $(a, b, c)$ are retained. Conversely, if fewer samples are needed, a subset is systematically selected. This approach enhances the minority class distribution, thereby addressing imbalance while preserving the underlying data manifold. The synthetic generation process ensures that oversampling occurs within the feature space boundaries of the minority class, reducing the risk of overfitting compared to naive duplication methods.

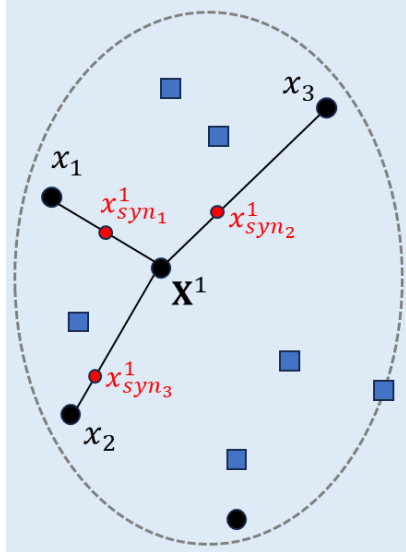

*Figure 3 An illustration of SMOTE algorithm. Consider an actual sample $X^1$, three nearest neighbors $x_1$, $x_2$ and $x_3$ are selected for linear calculation. Then three new synthetic points $x^1_{syn_1}$, $x^1_{syn_2}$ and $x^1_{syn_3}$ may occur as synthetic samples.*

A critical distinction lies in the application of SMOTE for minority class augmentation, which operates independently of majority class samples. This sets SMOTE as a data generation framework which is capable of synthesizing artificial instances for any targeted class group rather than being restricted to minority or majority subsets. In scenarios involving $N$ distinct classes, SMOTE can be deployed autonomously across each class, effectively serving as $N$ independent generators (from Generator 1 to Generator $N$). The resultant synthetic outputs, termed Pseudo-Class 1 to Pseudo-Class $N$ - are subsequently utilized. A multi-class anomaly detector will be trained with pseudo classes, which is further introduced in Sec. 3.4. It is important to note that, the number of anomaly categories does not need to be determined in advance. As new anomaly categories come in, the anomaly detection procedure could be gradually updated from binary-class detection, three-class detection, four-class detection and so on. Hence, to ensure that the detection model could identify normal signals and all known abnormal types, pseudo classes are essential to train the anomaly detection model, i.e., applying memory-replay strategies for continual learning. Algorithm 2 formalizes the integration of SMOTE within a replay-based continual learning framework. For each instance $x_i$ in class $i$, the algorithm first identifies its $k$-nearest neighbors within the class-specific feature space. Synthetic instances are generated through linear interpolation between $x_i$ and each neighbor, computed as randomly sampled points along the line segments connecting these pairs. The quantity of synthetic data is governed by predetermined augmentation criteria: if the target matches the cardinality of the original class, a single interpolated point is randomly selected per neighbor pair; otherwise, a proportional subset is retained. These synthesized instances are aggregated into a composite pseudo-class $i$, designed to emulate the feature distribution of the original class while addressing data scarcity. This pseudo-class is subsequently integrated into the continual learning pipeline to enhance classifier robustness against catastrophic forgetting, with implementation specifics elaborated in Section 3.4.

---

**Algorithm 2** SMOTE algorithm

---

**Input**: Samples in each class $\mathbf{X}_1,\ldots,\mathbf{X}_N$ from Algorithm 1; Number of nearest neighbors $k$; Number of generated samples $S$
**For** class $i$ **from** 1 **to** $N$:
   **Step 1:** Denote number of samples in $\mathbf{X}_i$ as $M$
   **For** sample $\mathbf{X}_i^j$ from 1 to $M$:
     **Step 2:** Calculate $k$ nearest neighbors $x_1,\ldots,x_k$ to $\mathbf{X}_i^j$
     **For** $x_l$ from 1 to k:
       **Step 3:** Obtain one random point as $\mathbf{x}_{syn_{l_i}}^j$ along the line segment between $\mathbf{X}_i^j$ and $x_l$
     **Step 4:** Choose $[\frac{S}{N}]$ random points as $\mathbf{x}_{syn_i}^j$ among $\{\mathbf{x}_{syn_{1_i}}^j, \ldots, \mathbf{x}_{syn_{k_i}}^j\}$
   **Step 5:** Combine $\{\mathbf{x}_{syn_i}^1, \ldots, \mathbf{x}_{syn_i}^M\}$ as pseudo class $i$
**Output:** Pseudo class 1, ..., Pseudo class $N$

---

## 3.4 Unsupervised Identification and Replay-based framework for ECG Anomaly Detection

This section delineates a generalized Unsupervised Identification and Replay-based Detection (UIRD) framework tailored for sequential process ECG signal anomaly detection, as illustrated in Figure 4. The methodology includes several tasks in sequence. It assumes an initial phase of nominal ECG signals, during which baseline data is collected and utilized to train Generator 1. This generator employs SMOTE to synthesize pseudo-samples that replicate the feature distribution of normal ECG signals. Upon the emergence of the first new type of ECG signals, i.e., Type 1 ECG anomaly signals, MadeGAN is applied to detect occurrence of such Type 1 ECG signals. Afterwards, a classifier is trained using a hybrid dataset comprising real Type 1 ECG signals and synthetic normal ECG signals generated by Generator 1, enabling discrimination between both types of ECG signals. Concurrently, MadeGAN is re-trained by both types of ECG signals to better learn further new types of ECG signals.

In addition, Generator 2 is trained with detected Type 1 ECG anomaly signals. When the second new type of ECG signals, i.e., Type 2 ECG anomaly signals, come in, the re-trained MadeGAN is applied to detect such Type 2 ECG anomaly signals. Then both the ECG signals from Generator 1 and Generator 2, combined with real Type 2 ECG anomaly signals, train a subsequent classifier to differentiate between all three types of ECG signals. Besides, MadeGAN is re-trained by such three types of ECG signals and Generator 3 is deployed to model the unique characteristics of Type 2 ECG signals. In this way, MadeGAN and generators are adaptively updated and the entire framework permits extensibility to additional types of ECG signals, ensuring scalable adaptation to evolving monitoring requirements of ECG signals while mitigating catastrophic forgetting through synthetic memory replay.

More importantly, in real-world scenarios, the decision rule for when to update the model is controlled by the MadeGAN component. When MadeGAN recognizes the input ECG anomaly signals as the signals from a new anomaly category, the entire model will be updated. That is, the overall procedure will be passed to next sequential task, MadeGAN and classifier will be updated, and a new SMOTE will be trained to generate pseudo ECG signals which belong to a new abnormal category.

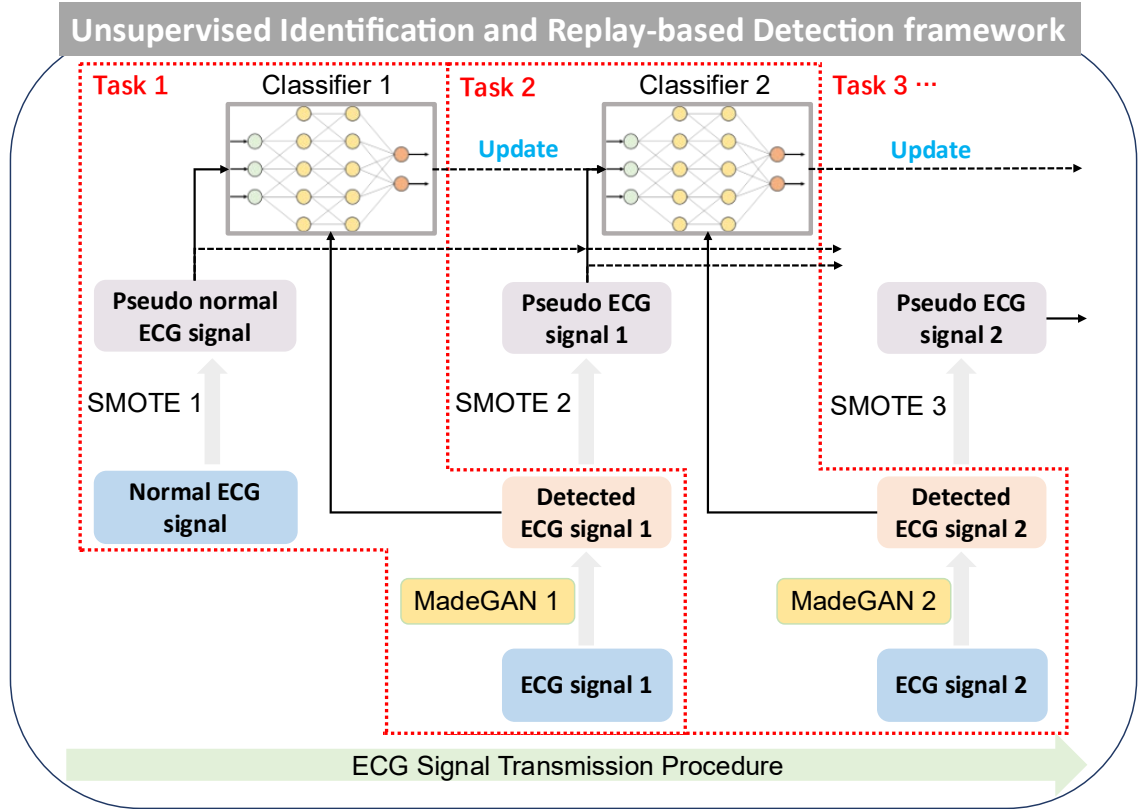

*Figure 4 Unsupervised identification and replay-based detection (UIRD) framework to detect and identify ECG anomaly signals sequentially.*

The overall framework is demonstrated in Algorithm 3. Specifically, when each new class of ECG anomaly signals come in, four steps are illustrated: (1) Detect such new class of ECG anomaly signals and save the detected new class ECG anomaly signals; (2) Generate pseudo data from previous classes and train the classifier based on pseudo data and the detected new class ECG anomaly signals; (3) Re-train MadeGAN based on detected new class ECG signals to further identify future new class of ECG anomaly signals; (4) Train a new generator based on detected new class ECG signals to generate pseudo data for next loop. In this way, the proposed method could be able to distinguish all previously observed ECG signals and identify newly unseen ECG signals.

---

**Algorithm 3** Replay-based Continual Learning for Multiple types of ECG signals

**Input**: Samples in normal ECG signals $\mathbf{X}_{ori_0}$ and different new types of ECG anomaly signals $\mathbf{X}_{ori_1},\ldots,\mathbf{X}_{ori_N}$;

**Step 1:** Train generator $G_0$ and MadeGAN $M_0$ for normal ECG signals $X_{ori_0}$
**For** class $i$ **from** 1 **to** $N$:
   **Step 2:** Detect $X_{ori_i}$ from $M_{i-1}$, and denote the detected ECG signals from the $i$-th type as $X_i$
   **Step 3:** Generate pseudo data by generator $G_0,\ldots, G_{i-1}$ representing existing classes $\widetilde{X}_0,\ldots,\widetilde{X}_{i-1}$;
   **Step 4:** Train classifier $C_i$ using $\{\widetilde{X}_0,\ldots,\widetilde{X}_{i-1}, X_i\}$
   **Step 5:** Re-train $M_i$ from $M_{i-1}$ based on $X_i$ from Algorithm 1
   **Step 6:** Train generator $G_i$ based on $X_i$ from Algorithm 2
   **Return** Classifier $C_i$, MadeGAN $M_i$ and generator $G_i$
**Output:** Classifier $C_N$, MadeGAN $M_N$ and generator $G_N$

It is important to note that the proposed UIRD framework exhibits distinct advantages in addressing continual learning challenges within ECG signals. Compared with other continual learning methods, the proposed framework embeds the unsupervised anomaly detection part and not only learns novel anomaly incrementally but is also able to detect the occurrence of new anomaly patterns. Empirical evidence suggests that replay-based methodologies outperform alternative continual learning paradigms in class-incremental continual learning scenarios, demonstrating superior efficacy in preserving knowledge across sequential tasks - a critical requirement for dynamic ECG detection. Furthermore, the integration of pseudo-data generation circumvents computational and storage bottlenecks inherent to large-scale data retention, as the framework requires only a subset of representative ECG signals for synthetic data synthesis which are detected from MadeGAN. Notably, the adoption of SMOTE as the pseudo-data generator enhances feature-space representativeness by synthesizing high-fidelity pseudo-samples that attenuate outlier influence and enrich the training manifold, thereby potentially achieve comparable monitoring performance with raw data alone. To validate these theoretical advantages, Sec. 4 presents four industrial case studies of ECG signals designed to empirically evaluate the framework's operational efficacy and generalizability in real-world healthcare contexts.

## 4 Real-world case study

To ascertain the efficacy of the proposed methodology, this section undertakes an evaluation of its performance within four different real-world datasets. The overall dataset introduction is demonstrated in Sec 4.1. Afterwards, the comprehensive experiments of the proposed method on MITDB dataset are demonstrated in Sec. 4.2 while Sec. 4.3 discusses the results of the proposed method on all the other three datasets.

## 4.1 Dataset introduction

To demonstrate the effectiveness of the proposed method effectively, four well-established datasets are applied in this paper as follows:

1. **MIT-BIH Arrhythmia Database (MITDB)** [35] - Contains 48 half-hour two-lead ambulatory ECG recordings from 47 individuals, and is one of the most widely used benchmarks for arrhythmia detection algorithms. It includes annotations for six beat types (normal and arrhythmic) and has been the basis for countless studies in arrhythmia classification.

2. **MIT-BIH Supraventricular Arrhythmia Database (SVDB)** [36] - Contains 78 ECG recordings (30 minutes each) that supplement MITDB with more examples of supraventricular arrhythmias (e.g., premature atrial beats). This dataset specifically addresses arrhythmias originating above the ventricles, providing additional data for those less-common rhythm disorders.

3. **MIT-BIH Noise Stress Test Database (NSTDB)** [37] - Consists of 12 ECG recordings (30 minutes each) that have segments of synthetic noise (baseline wander, muscle artifact, electrode motion) artificially added to clean ECG signals. It is used to assess how algorithms perform under noisy conditions, ensuring robustness of arrhythmia detectors in real-world ambulatory settings.

4. **European ST-T Database (EDB)** [38] - Contains 90 two-hour ECG excerpts from patients, annotated with episodes of ST segment and T-wave abnormalities (e.g., ischemic changes). This database is a standard for evaluating algorithms that detect ischemia or other ST-T deviations, complementing arrhythmia-focused datasets with a focus on morphological changes in the ECG waveform.

To demonstrate the performance of the proposed method across multiple heartbeat categories, we design a series of experiments structured into sequential tasks. Across all ECG datasets used in this study, we focus on six distinct heartbeat types: Normal (N), Left bundle branch block (L), Right bundle branch block (R), Premature ventricular contraction (V), Atrial premature beat (A), and Fusion of ventricular and normal beat (f). Because most public ECG datasets do not contain sufficient samples for all categories—particularly for rare classes such as fusion beats—we restricted our incremental learning setup to six representative heartbeat classes. This choice ensures both clinical relevance and adequate data support for each class, avoiding extreme imbalance that could undermine fair evaluation. Specifically, the experiment is divided into five sequential tasks (Task 1 to Task 5 as shown below). Each task introduces a new heartbeat anomaly type that the model has not previously encountered. The model is required to classify all six heartbeat types throughout the experiment, where each task introduces a new ECG class. It is important to note that, the class introduction order of tasks could be randomized. To make the class-introduction order in different cases more clear and consistent, different classes are sent into the model based on its sample size in this study, starting with the most common—normal heartbeats.

- **Task 1:** Augment pseudo samples of "N" ECG signals, distinguish "L" ECG signals and "N" ECG signals
- **Task 2:** Augment pseudo samples of detected "L" ECG signals. When the third type of ECG signals, i.e., "R" ECG signals, comes in, classify all three real-world ECG signals from the pseudo "N" and "L" signals and the collected "R" ECG signals
- **Task 3:** Augment pseudo samples of detected "R" ECG signals. When "V" ECG signals come in, classify all four ECG signals from the pseudo signals and the collected "R" ECG signals
- **Task 4:** Augment pseudo samples of detected "V" ECG signals. When "A" ECG signals come in, classify all five ECG signals from the pseudo signals and the collected "A" ECG signals
- **Task 5:** Augment pseudo samples of detected "A" ECG signals. When "f" ECG signals, come in, classify all six ECG signals from the pseudo signals and collected "f" ECG signals

Each heartbeat is extracted using a multi-step process: (1) R-peaks are identified from the raw ECG recordings using the Pan-Tompkins QRS detection algorithm [39]; (2) a fixed-length window of 320 samples is centered around each R-peak, with 160 samples preceding and 160 samples following the peak. This ensures uniform signal length and coverage of a full cardiac cycle. To mitigate common ECG noise, such as baseline drift and power-line interference, a high-pass finite impulse response (FIR) filter is applied. The classification metrics, including precision, recall and F-scores are calculated to demonstrate the effectiveness of the proposed method. Furthermore, comparative analysis is conducted by implementing several established benchmark approaches. Specifically, the continual learning method, Elastic Weight Consolidation (EWC), is adopted as a representative continual learning strategy. In this study, a Convolution Neural Network (CNN) classifier comprising two 1-D convolutional layer and three fully connected layers is employed as the base classifier. To ensure consistency and fairness in the evaluation, all approaches considered in this work utilize the same architecture. Additionally, standardization is applied to all input samples to maintain compatibility with the neural network structure and to enhance model performance.

In addition to EWC, the anomaly detection experiments based on basic MadeGAN approach is also included as ablation experiments to demonstrate effectiveness of continual learning framework. This method adopts the same CNN model structure and operates by initially training the model on both normal and abnormal data from Task 1. Subsequently, the model is further trained using normal data and detect newly introduced abnormal data from Task 2, after which it is evaluated on Task 2. The other tasks follow the same setup.

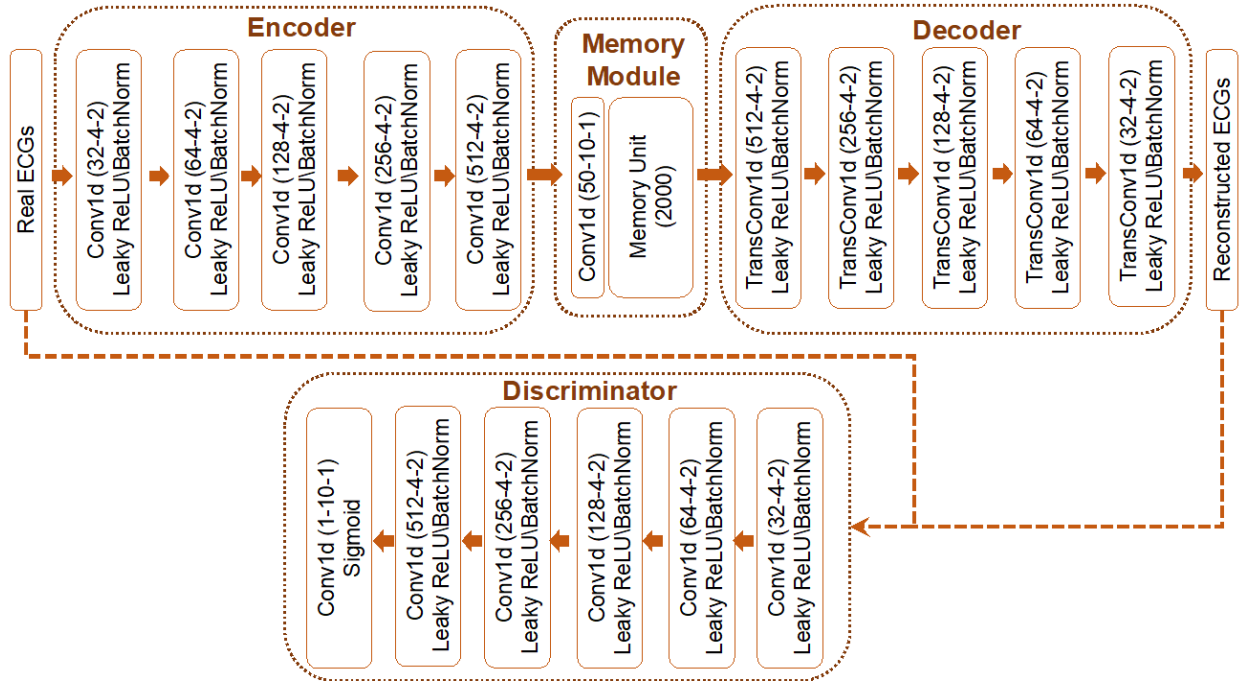

*Figure 5 The architecture details of MadeGAN framework.*

The Figure 5. illustrates the architecture of MadeGAN. Specifically, the MadeGAN architecture is composed of four major components: an encoder, a decoder, a memory module, and a discriminator. The encoder consists of a sequence of 1D convolutional layers with progressively increasing channel sizes, each followed by LeakyReLU activation and batch normalization. Note that the notation of (512-4-2), for example, means that this 1D CNN layer consists of 512 filters with the filter size of 4 and stride of 2. This stack encodes the input ECG signal into a compact latent representation. The memory unit has fixed-size slots 2000 (i.e., prototype vectors). The decoder mirrors the encoder but uses transposed convolutional layers (TransConv1d) to reconstruct the input signal from the memory-augmented latent space. The discriminator receives both real ECGs and reconstructed ECGs. It is built with a series of 1D convolutional layers (Conv1d), LeakyReLU activations, and batch normalization, similar to the encoder.

Specifically, the only differences between the basic MadeGAN approach and the proposed method are the input samples, which are actual samples and pseudo samples, separately. Hence, if the proposed method could achieve the same performance with the basic MadeGAN approach, the quality pseudo samples should be the same as the actual samples.

For comprehensive comparison, a baseline approach is also incorporated, representing a classification scenario without any continual learning techniques. In the $i$-th task of this case, the model is trained using data from all $i$ classes simultaneously, providing an upper-bound performance reference for the other approaches. Compared with other benchmark methods, the samples utilized in the baseline approach are all real and complete. Hence, theoretically, the classification performance of baseline should be the supremum of the classification performance of all the other approaches.

Since MITDB includes the largest number of classes, the results under MITDB dataset are priorly demonstrated in Sec. 4.2 to show the effectiveness of applying MadeGAN as the base of anomaly detection and the superior performance of the proposed UIRD method. Afterwards, the results of all the other three datasets are discussed in Sec. 4.3.

### 4.2 Results and discussion based on MITDB dataset

In this section, the results of the proposed method on MITDB dataset are discussed. Since the proposed method includes two steps: anomaly detection and continual learning, the effectiveness of both two steps needs to be validated. Hence, the performance of anomaly detection is initially investigated in Sec. 4.2.1. Then the effectiveness of continual learning and the entire framework is demonstrated in Sec. 4.2.2.

#### 4.2.1 Anomaly detection performance on MITDB dataset

In this study, 80% samples are applied as the training set while 20% samples are applied as the testing set. Table 1 presents a comparative evaluation on the test dataset of MadeGAN against three benchmark models, MemAE [40], BeatGAN [41], and traditional autoencoder (AE) on five sequential classification tasks in an evolving ECG anomaly detection setting similar to the five tasks in Sec. 4.1. Each task is to distinguish a newly introduced ECG class (the "novel" class) from previously known samples of existing classes. Following the same class order in Sec. 4.1, the first task involves distinguishing L-type ECG signals from N-type signals. Then Subsequent tasks follow a continual learning setup where the known classes expand: the second task distinguishes R-type signals from N and L signals; the third separates V-type signals from N, L, and R signals; the fourth isolates A-type signals from N, L, R, and V signals; and the final task classifies f-type signals against N, L, R, V, and A signals.

As shown in Table 1, across all five tasks, MadeGAN consistently achieves the highest or competitive F-scores, indicating strong detection performance as new classes are introduced. In the first three settings, MadeGAN achieves F-scores of 0.92, 0.85, and 0.90 respectively, outperforming or matching the baselines. This shows its strength in early-stage binary classification with relatively simpler decision boundaries. In more challenging tasks, particularly those involving A and f signals where the normal class becomes more complex and diverse, MadeGAN maintains its superiority with F-scores of 0.64 and 0.49. Its precisions in these later tasks also surpass other methods, suggesting better reliability and fewer false alarms under increasing class imbalance and data complexity. Overall, the results highlight MadeGAN's robustness and generalization ability in continual ECG anomaly detection, where the decision boundary must adapt dynamically as the "normal" class evolves over time.

**TABLE 1:** *The comparison of average precisions, recalls and F-scores between MadeGAN and other anomaly detection benchmarks from Task 1 to Task 5*

| Method | Task 1 | | | Task 2 | | | Task 3 | | |
|---|---|---|---|---|---|---|---|---|---|
| | Precision | Recall | F-score | Precision | Recall | F-score | Precision | Recall | F-score |
| **MadeGAN** | **0.88** | **0.96** | **0.92** | **0.79** | **0.92** | **0.85** | **0.88** | **0.93** | **0.90** |
| MemAE | 0.88 | 0.97 | 0.92 | 0.73 | 0.95 | 0.83 | 0.79 | 0.89 | 0.84 |
| BeatGAN | 0.80 | 0,.92 | 0.86 | 0.71 | 0.96 | 0.82 | 0.86 | 0.85 | 0.86 |
| AE | 0.73 | 0.95 | 0.83 | 0.73 | 0.89 | 0.80 | 0.88 | 0.84 | 0.86 |

| Method | Task 4 | | | Task 5 | | |
|---|---|---|---|---|---|---|
| | Precision | Recall | F-score | Precision | Recall | F-score |
| **MadeGAN** | **0.52** | **0.85** | **0.64** | **0.38** | **0.71** | **0.49** |
| MemAE | 0.30 | 0.86 | 0.45 | 0.29 | 0.60 | 0.40 |
| BeatGAN | 0.49 | 0.84 | 0.62 | 0.25 | 0.68 | 0.36 |
| AE | 0.34 | 0.89 | 0.49 | 0.31 | 0.62 | 0.41 |

### 4.2.2 Unsupervised identification and replay-based detection performance on MITDB dataset

By applying MadeGAN as the base of anomaly detection part, the average precisions, recalls and F-scores of all the approaches for all 5 tasks are shown in Table 2. It is important to highlight that, due to the shared initial stage among all approaches—namely, the training of a binary classifier—the F-scores for Task 1 are predominantly identical, with most approaches achieving a value of 0.99.

*TABLE 2 : Average precisions, recalls and F-scores from Task 1 to Task 5*

| Method | Task 1 | | | Task 2 | | | Task 3 | | |
|---|---|---|---|---|---|---|---|---|---|
| | Precision | Recall | F-score | Precision | Recall | F-score | Precision | Recall | F-score |
| EWC | 0.99 | 0.99 | 0.99 | 0.44 | 0.66 | 0.50 | 0.29 | 0.49 | 0.32 |
| MadeGAN | 0.99 | 1.00 | 0.99 | 0.94 | 0.68 | 0.77 | 0.89 | 0.61 | 0.71 |
| Baseline | 0.99 | 1.00 | 0.99 | 0.88 | 0.80 | 0.81 | 0.83 | 0.69 | 0.71 |
| **UIRD (Proposed)** | **0.99** | **0.99** | **0.99** | **0.86** | **0.81** | **0.82** | **0.81** | **0.70** | **0.73** |

| Method | Task 4 | | | Task 5 | | |
|---|---|---|---|---|---|---|
| | Precision | Recall | F-score | Precision | Recall | F-score |
| EWC | 0.35 | 0.40 | 0.28 | 0.19 | 0.33 | 0.21 |
| MadeGAN | 0.74 | 0.50 | 0.57 | 0.61 | 0.42 | 0.46 |
| Baseline | 0.66 | 0.56 | 0.57 | 0.57 | 0.48 | 0.48 |
| **UIRD (Proposed)** | **0.63** | **0.58** | **0.59** | **0.56** | **0.48** | **0.47** |

In all Task 2, Task 3 and Task 4, the proposed method achieves the highest recall and F-score among all evaluated benchmark approaches, demonstrating its superior capability in detecting both previously seen and newly introduced ECG signals. Specifically, the relatively lower precision in all the tasks of the proposed method is due to the changes of sample size. The second type of ECG signals does not have the second largest number of detected samples. Hence, the pseudo samples are not effective enough to guarantee the same precision as other approaches in Task 2-5. Even though the proposed method does not have the best F-score in Task 5, it is still comparable to the baseline method.

It is shown that the performance of MadeGAN in Table 1 is different from the performance in Table 2, which is due to the sample size. When performing the sequential tasks shown in Table 2, the samples from each type of ECG signals are selected according to the MadeGAN performance from the previous tasks.

Under such circumstance, each type of ECG signals has less samples to train MadeGAN in the tasks shown in Table 2 than the tasks shown in Table 1. Hence, the performance of MadeGAN in Table 1 is better than the performance in Table 2.

In addition, continual learning-based methods typically struggle to outperform the baseline approach, as evidenced by the results presented in previous tables. However, the proposed method notably surpasses the baseline in terms of F-score in several tasks, suggesting that the integration of SMOTE not only facilitates the generation of representative samples for each class but also contributes to reducing intra-class variance and enhancing inter-class separability. This, in turn, enables the proposed method to achieve superior performance in the class-incremental continual learning scenario, a result unattainable by other benchmark approaches. These findings suggest that, despite potential limitations in data quality, the proposed method remains effective in producing representative samples that can significantly enhance classification performance.

Furthermore, applying CNN as the classifier in each task is due to excellent classification performance of CNN. However, it is important to note that, the proposed replay-based approach provides greater flexibility. That is, unlike regularization-based continual learning methods, which are constrained by fixed model architectures, when performing replay-based detection, different classifiers could be applied in different tasks according to their performance. This allows for the use of different model architectures tailored to the characteristics of the data, optimizing model selection without architectural restrictions.

Since the anomaly categories are sent into the model in sequential, catastrophic forgetting may occur, which means the performance of the model when identifying the samples from previous tasks may significantly decrease. Hence, to further validate the performance of the proposed method under old tasks, the F-scores of all the classes under different tasks are demonstrated in Table 3. the F-scores of EWC, MadeGAN and baseline are also illustrated in this table for comparison.

***TABLE 3 :*** *Average F-scores of each class under different tasks*

| Method | "N" type | "L" type |
|---|---|---|

| Method | "N" type | | | | | "L" type | | | | |
|---|---|---|---|---|---|---|---|---|---|---|
| | Task 1 | Task 2 | Task 3 | Task 4 | Task 5 | Task 1 | Task 2 | Task 3 | Task 4 | Task 5 |
| EWC | 1,00 | 1.00 | 1.00 | 1.00 | 1.00 | 0.98 | 0.00 | 0.00 | 0.00 | 0.00 |
| MadeGAN | 1,00 | 0.97 | 0.97 | 0.95 | 0.94 | 0.98 | 0.72 | 0.71 | 0.68 | 0.65 |
| Baseline | 1,00 | 1.00 | 1.00 | 1.00 | 0.99 | 0.98 | 0.85 | 0.80 | 0.69 | 0.63 |
| **UIRD (Proposed)** | 1,00 | 1.00 | 1.00 | 1.00 | 1.00 | **0.99** | **0.85** | **0.81** | **0.70** | **0.65** |

| Method | "R" type | | | | | "V" type | | | | |
|---|---|---|---|---|---|---|---|---|---|---|
| | Task 1 | Task 2 | Task 3 | Task 4 | Task 5 | Task 1 | Task 2 | Task 3 | Task 4 | Task 5 |
| EWC | — | 0.50 | 0.00 | 0.00 | 0.00 | — | — | 0.28 | 0.01 | 0.00 |
| MadeGAN | — | 0.61 | 0.57 | 0.53 | 0.49 | — | — | **0.57** | **0.42** | 0.35 |
| Baseline | — | 0.59 | 0.54 | 0.50 | 0.45 | — | — | 0.51 | 0.32 | **0.36** |
| **UIRD (Proposed)** | — | **0.61** | **0.58** | **0.53** | **0.50** | — | — | 0.52 | 0.37 | 0.27 |

| Method | "A" type | | | | | "f" type | | | | |
|---|---|---|---|---|---|---|---|---|---|---|
| | Task 1 | Task 2 | Task 3 | Task 4 | Task 5 | Task 1 | Task 2 | Task 3 | Task 4 | Task 5 |
| EWC | — | — | — | **0.41** | 0.00 | — | — | — | — | **0.23** |
| MadeGAN | — | — | — | 0.27 | 0.26 | — | — | — | — | 0.09 |
| Baseline | — | — | — | 0.34 | 0.30 | — | — | — | — | 0.15 |
| **UIRD (Proposed)** | — | — | — | 0.34 | **0.31** | — | — | — | — | 0.11 |

As the number of tasks increases, the F-scores of existing ECG types decrease due to the catastrophic forgetting. However, due to the generation of pseudo classes, the proposed method could train the classifiers with comprehensive pseudo samples, leading to the mostly highest F-scores of each class under different tasks, especially for the relatively old classes, like "N", "L" and "R" types of ECG signals. Besides, in each task, the overall F-score performance of the proposed method, as shown in Table 2, is the best among all the approaches. Hence, it also demonstrates the relatively strong ability to handle the catastrophic forgetting issue of the proposed UIRD framework. Overall, these results provide strong empirical evidence supporting the effectiveness and robustness of the proposed UIRD framework in class-incremental continual learning tasks. The performance of the proposed method is also validated based on the other three datasets, which are illustrated in Sec. 4.3.

## 4.3 Results and discussion based on SVDB, NSTDB and EDB datasets

### 4.3.1 SVDB dataset-based case study

This sub-section illustrates the performance of the proposed method based on SVDB dataset. Similarly, we assume that these three types of ECG signals manifest sequentially according to the sample size, encompassing Task 1 and Task 2 as delineated below:

- **Task 1:** Augment pseudo samples of "N" ECG signals, distinguish "S" ECG signals and "N" ECG signals

- **Task 2:** Augment pseudo samples of detected "S" ECG signals. When the third type of ECG signals, i.e., "V" ECG signals, comes in, classify all three ECG signals from the collected samples

The other experimental setups are the same as the setup described in Sec. 4.1. Table 4 presents the average precision, recall, and F-scores obtained by all evaluated approaches across both Task 1 and Task 2. It is worth noting that, due to the common initial stage in all methods — specifically, the training of a binary classifier — the F-scores for Task 1 are generally consistent, with most approaches achieving a score of approximately 0.96. An exception is observed in the case of the EWC method, which yields a lower F-score of 0.94, indicating slightly inferior performance relative to the other approaches.

*TABLE 4: Average precisions, recalls and F-scores of Task 1 and 2*

| Method | Task 1 | | | Task 2 | | |
|---|---|---|---|---|---|---|
| | Precision | Recall | F-score | Precision | Recall | F-score |
| EWC | 0.98 | 0.97 | 0.97 | 0.56 | 0.66 | 0.60 |
| MadeGAN | 0.97 | 0.96 | 0.96 | 0.75 | 0.72 | 0.73 |
| Baseline | 0.97 | 0.96 | 0.96 | 0.77 | 0.77 | 0.77 |
| **UIRD (Proposed)** | **0.95** | **0.96** | **0.96** | **0.76** | **0.73** | **0.74** |

In contrast, for Task 2, the proposed method achieves superior performance than the benchmark approaches in terms of precision, recall, and F-score. Though it is not better than the baseline, the results are still comparable. This result highlights the enhanced capability of the proposed method in accurately detecting abnormal ECG signals. It is important to emphasize that, as indicated in Table 4, continual learning-based methods typically struggle to outperform the baseline approach, which has access to all class data simultaneously. While the F-score achieved by the proposed method in Task 2 does not exceed that of the baseline, it remains equivalent, demonstrating competitive performance. In comparison, all other approaches exhibit substantially lower metric values.

### 4.3.2 NSTDB dataset-based case study

NSTDB dataset includes three classes so that the detect-incremental tasks are structured as two tasks. According to the sample size, the detection task between "N" ECG signals and pseudo "R" ECG signals is considered as Task 1, while the detection task between "V" ECG signals and "R", "N" ECG signals is considered as Task 2. In addition, all the other experimental setups are the same as the setup in Sec. 4.1.

Table 5 presents the average precision, recall, and F-scores for all evaluated approaches in both Task 1 and Task 2. All approaches, including the baseline, achieve a perfect score of 1 across all evaluation metrics in Task 1, indicating successful task completion. However, it is important to emphasize that such high classification metrics do not necessarily imply that all individual samples are correctly classified. In particular, it is observed that approximately 1 to 3 samples from Case 1 are consistently misclassified as belonging to the normal class. This observation suggests that, despite the overall high performance, there remains potential for improvement in accurately distinguishing samples from normal class.

*TABLE 5: Average precisions, recalls and F-scores of Task 1 and 2*

| Method | Task 1 | | | Task 2 | | |
| --- | --- | --- | --- | --- | --- | --- |
| | Precision | Recall | F-score | Precision | Recall | F-score |
| EWC | | | | 0.52 | 0.67 | 0.57 |
| MadeGAN | | | | 0.85 | 0.85 | 0.84 |
| Baseline | 1.00 | 1.00 | 1.00 | 0.82 | 0.82 | 0.82 |
| **UIRD (Proposed)** | | | | **0.86** | **0.86** | **0.86** |

In contrast, for Task 2, the proposed approach outperforms all benchmark methods in terms of precision, recall, and F-score, demonstrating enhanced effectiveness in detecting abnormal ECG signals. Notably, as indicated in Table 5, continual learning-based approaches conventionally exhibit inferior performance relative to the baseline. The F-scores of the proposed method are higher than that of the baseline, whereas all other approaches achieve substantially lower scores, suggesting the method's robustness and its ability to consistently generate representative data.

4.3.3 EDB dataset-based case study

In this section, EDB dataset is applied to demonstrate the performance of the proposed method under four classes. According to the sample size, there will be three tasks as illustrated below:

- **Task 1:** Augment pseudo samples of "V" ECG signals, distinguish "T" ECG signals and "V" ECG signals
- **Task 2:** Augment pseudo samples of detected "T" ECG signals. When the third type of ECG signals, i.e., "S" ECG signals, comes in, classify all three ECG signals from the collected samples
- **Task 3:** Augment pseudo samples of detected "S" ECG signals. When the fourth type of ECG signals, i.e., "F" ECG signals, comes in, classify all four ECG signals from the collected samples

The detailed metrics of all three tasks based on EDB dataset are provided in Table 6. Overall, the proposed method mostly achieves the highest values across most classification metrics in Task 1, which demonstrates the effectiveness of the proposed method as the number of actual samples is large.

*TABLE 6: Average precisions, recalls and F-scores from Task 1 to Task 3*

| Method | Task 1 | | | Task 2 | | | Task 3 | | |
|---|---|---|---|---|---|---|---|---|---|
| | Precision | Recall | F-score | Precision | Recall | F-score | Precision | Recall | F-score |
| EWC | 0.96 | 0.92 | 0.94 | 0.58 | 0.63 | 0.60 | 0.35 | 0.49 | 0.39 |
| MadeGAN | 0.90 | 0.95 | 0.92 | 0.77 | 0.69 | 0.70 | 0.56 | 0.53 | 0.52 |
| Baseline | 0.90 | 0.95 | 0.92 | 0.75 | 0.70 | 0.71 | 0.56 | 0.53 | 0.52 |
| **UIRD (Proposed)** | **0.92** | **0.92** | **0.92** | **0.76** | **0.70** | **0.69** | **0.55** | **0.54** | **0.53** |

However, the performance of the proposed method in Task 2 could not beat both MadeGAN and baseline, which is due to the sample selection. Though the different types of ECG signals are sent to the tasks following the order of sample size, the order of detected sample size may change. That is, the number of detected "T" ECG signals are the smallest in all the detected ECG signals. Then the proposed method does not have sufficient samples to guarantee the quality of pseudo "T" ECG signals. Afterwards, when "S" ECG signals are applied in Task 2, the performance of the proposed method becomes worse than the other approaches which use the actual samples.

Additionally, although the precision of the proposed method is occasionally lower than that of the baseline or MadeGAN, its recalls and F-scores remain superior in Task 3. This trend is particularly evident for the last one task introduced, where the proposed method consistently attains the highest F-scores among all evaluated approaches.

# 5   Conclusions and future work

This study introduces a novel Unsupervised Identification and Replay-based Detection (UIRD) framework, augmented with synthetic data generation, to address anomaly detection based on ECG signals. The framework aims to enhance the detection accuracy of all previously observed ECG signal patterns and identification accuracy of newly introduced ECG signals. Its contributions are threefold: First, it unifies anomaly detection with continual learning mechanisms, enabling dynamic model adaptation to emergent anomalies through real-time updates. Second, the pseudo-replay methodology optimizes the equilibrium between computational efficiency and diagnostic accuracy, rendering it applicable to resource-constrained industrial settings by generating and saving pseudo samples. Third, the framework introduces architectural plasticity, permitting the deployment of heterogeneous model configurations that adaptively align with evolving data structures and anomaly signatures.

Empirical validation via case studies on four ECG signal datasets demonstrates the framework's superiority, with UIRD achieving consistently comparable precision, recall, and F1-scores compared to baselines, highlighting its efficacy in sustaining and augmenting monitoring performance. Future investigations will explore the impact of alternative generative algorithms on framework performance and extend validation to diverse healthcare domains, broadening adoptability to complex real-world applications.